\documentclass[a4paper,twoside]{article}

\usepackage[utf8]{inputenc} % allow utf-8 input
\usepackage[T1]{fontenc}    % use 8-bit T1 fonts

\usepackage{graphicx}
\usepackage{hyperref}
\usepackage{mathtools}
\usepackage{booktabs}
\usepackage{multirow}
\usepackage{multicol}
\usepackage{siunitx}

\usepackage{epsfig}
\usepackage{subcaption}
\usepackage{calc}
\usepackage{amssymb}
\usepackage{amstext}
\usepackage{amsmath}
\usepackage{amsthm}
\usepackage{multicol}
\usepackage{pslatex}
\usepackage{apalike}

\usepackage{SCITEPRESS}

\hypersetup{
	pdftitle={Representational Capacity of Deep Neural Networks -- A Computing Study},
	pdfauthor={Bernhard Bermeitinger, Tomas Hrycej, Siegfried Handschuh},
	pdfkeywords={deep learning, artificial neural networks, optimization, conjugate gradient}
}

\begin{document}

\title{Representational Capacity of Deep Neural Networks -- A Computing Study}

\author{\authorname{Bernhard Bermeitinger\sup{1,2}\orcidAuthor{0000-0002-2524-1850}, Tomas Hrycej\sup{1} and Siegfried Handschuh\sup{1,2}}
	\affiliation{\sup{1}Institute for Computer Science, University of St.Gallen, St.Gallen, Switzerland}
	\affiliation{\sup{2}Faculty for Computer Science, University Passau, Passau, Germany}
	\email{\{bernhard.bermeitinger, tomas.hrycej, siegfried.handschuh\}@unisg.ch, \{siegfried.handschuh\}@uni-passau.de}
}


\abstract{
	There is some theoretical evidence that deep neural networks with multiple hidden layers have a potential for more efficient representation of multidimensional mappings than shallow networks with a single hidden layer. The question is whether it is possible to exploit this theoretical advantage for finding such representations with help of numerical training methods. Tests using prototypical problems with a known mean square minimum did not confirm this hypothesis. Minima found with the help of deep networks have always been worse than those found using shallow networks. This does not directly contradict the theoretical findings---it is possible that the superior representational capacity of deep networks is genuine while finding the mean square minimum of such deep networks is a substantially harder problem than with shallow ones.}

\onecolumn \maketitle \normalsize \setcounter{footnote}{0} \vfill

% !TeX encoding = UTF-8
% !TeX spellcheck = en_US
\section{\uppercase{Introduction}}\label{sec:introduction}
\noindent At present, there is a strong revival of interest in layered neural networks. Although the basic structures and algorithms are similar to those developed in the eighties of the past century, there are some shifts in the focus. The most important of them is the emphasis on large networks with more than one hidden layer. The current interest wave is motivated by numerous reports about positive computing experience with such multi-layer networks for very large mapping tasks. Typical applications are corpus-based semantics and computer vision. In particular the former is characterized by a strong under-determination of model parameters---there are substantially more parameters than labeled training examples.
There are some review works on deep learning, summarizing both the success stories and the theoretical justifications for the success. This paper takes the extensive work by Lecun~et~al.~\cite{lecun_DeepLearning_2015} as a frequent reference. 

The term \emph{deep networks} will be used in the sense of deep learning, denoting networks with more than one hidden layer. Networks with a single hidden layer will be referred to as \emph{shallow networks}.

Besides some experimental findings of deep network representational efficiency, there are also several attempts for theoretical justifications. 
They state essentially the following: deep networks exhibit larger representation capacity than shallow networks. That is, they are capable of approximating a broader class of functions with the same number of parameters.~\cite{montufar_NumberLinearRegionsDeepNeural_2014}

To compare the representation capacity of deep and shallow networks, several interesting results have been published. Bengio~et~al.~\cite{bengio_OutofsampleExtensionsLleIsomapMds_2004} have investigated a class of algebraic functions that can be represented by a special structure of deep networks (alternating summation and product layers). They showed that for a certain type of deep networks the number of hidden units necessary for a shallow representation would grow exponentially while in a deep network it grows only polynomial.
Montufar~et~al.~\cite{montufar_NumberLinearRegionsDeepNeural_2014} have used a different approach for evaluating the representational capacity. They investigated how many different linear hyperplane regions of input space can be mapped to an output unit. They derived statements for the maximum number of such hyperplanes in deep networks, showing that this number grows exponentially with the number of hidden layers. The activation units used have been rectified linear units and softmax units.
It is common to these findings that they do not make statements about arbitrary functions. The result of \cite{bengio_OutofsampleExtensionsLleIsomapMds_2004} is valid for algebraic functions representable by the given deep network architecture, but not for arbitrary algebraic terms. \cite{montufar_NumberLinearRegionsDeepNeural_2014} have derived the maximum number of representable hyperplanes, but it is not guaranteed, that this maximum can be attained for an arbitrary function to be represented. In other words, there are function classes that can be efficiently represented by a deep network, while other functions cannot.
This is not unexpected: knowing that some $ N_1 $ dimensional function is to be identified from $ N_2 $ training examples (which is equivalent to satisfying $ N = N_1 \times N_2 $ equations), it cannot be generally represented by less than $ N $ parameters although cases representable by fewer parameters exist. Another familiar analogy is that of algebraic terms. Some of them can be made compact by the distributive law, others cannot.

This finding can be summarized in the following way. There exist mappings that can be represented by deep neural networks more economically than by shallow ones. These mappings are characterized by multiple usages of intermediary (or hidden) concepts. This may be typical for cognitive mappings, so that deep networks may be adequate for cognitive tasks.

There are various studies claiming the superiority of deep networks based on positive computing experience. 
Most of them concern particular architectures such as networks using convolutional layers (e.g.,~\cite{goodfellow_MultidigitNumberRecognitionStreetView_2013}). 
This network type has strong justification for image recognition since the convolutional layers are closely related to spacial operators known to be important for image processing. 
It is then logical to expect that a network with an appropriate number of such layers is superior to a shallow network that offers only the possibility of using a single convolutional layer followed directly by the final processing to the output.

A few works address the issue of the representational capacity of fully connected deep networks. 
\cite{erhan_DifficultyTrainingDeepArchitecturesEffect_2009} is aware of problems with convergence properties of optimization algorithms for deep networks, but claim their superiority to shallow networks on test sets. 
Their particular focus is to show the usefulness of unsupervised pre-training of some layers, rather than the comparison of the representational capacity so that the choice of test problems makes it not fully appropriate for clarifying the representational capacity of deep and shallow networks.
\begin{itemize}
    \item Their study compares networks with different total numbers of network parameters (weights and biases) 
	so that the comparison is favorable for deeper networks having more parameters.
    \item Simple first order optimization methods with fixed learning rates are used so that the danger of influencing the results by a poor convergence of the optimization is serious.
    \item Some problems are under-determined, that is, having fewer constraints than parameters. In this case, the minimum of the training set error may be expected to be zero not only for a single optimum solution but also for large subsets of the parameter space.
\end{itemize}

For an objective review of deep and shallow networks, it is important to
\begin{itemize}
    \item use sufficiently determined problems (i.e., having more constraints than parameters)
    \item use optimization methods that can be expected to reliably converge at least to local minima
    \item compare shallow and deep networks with identical or nearly identical numbers of free network parameters
\end{itemize}
A comparison following these principles is the goal of this study.

\section{\uppercase{Attainable Representational Capacity}}\label{sec:representation-capacity}
\noindent Even if some class of functions has a superior representational capacity, it is not guaranteed that this capacity can be fully exploited---the additional necessary component is an algorithm that is capable of attaining the functional fit.

In terms of shallow and deep networks, it would be necessary to have algorithms that exploit the assumed superior capacity of deep networks in fitting them to a set of training examples in an efficient way. 
This efficiency would have to be sufficient not to lose the representational advantage. 
In practical terms, the usefulness of a network type consists of both a representational capacity and the algorithm efficiency. 
So, a high capacity potential and a poorly converging algorithm may result in low exploitable capacity.

In the concrete case of mean square minimization, the numerical efficiency of the optimizing algorithm for a given problem is the key parameter. For strictly convex minimization tasks, the usual measure of the potential efficiency is the condition number of the Hessian matrix (i.e., the matrix of the second derivatives of the error function with regard to the network parameters). 
This condition number is defined as the ratio of the largest and the smallest eigenvalue of the Hessian. 
Unfortunately, this objective measure cannot be applied for our comparison, for two reasons:
\begin{itemize}
    \item The error function is not convex at points far away from the minimum.
    \item Even at the points where the error function is convex, some eigenvalues are very close to zero, corresponding to search directions which have no or very small effect on the error function. 
    The condition number is then near to infinity. These directions result from redundancies inherent to the neural networks. This is the case, for example, due to the rotational invariance of the hidden layers at the points of nearly linear activation. Then, the mapping $ W_{i+1} W_i $ is identical with $ W_{i+1} H^{-1} H W_i $, for any non-singular square matrix $H$ of the dimension corresponding to the width of the $ i $-th hidden layer. So, the neural network constitutes the same mapping with the matrices $ W_i $ and $ W_{i+1} $ as with $ H W_i $ and $ W_{i+1} H^{-1} $.
\end{itemize}

For the lack of theoretical alternatives, the only possibility is to assess the efficiency of particular problems experimentally.
To make reliable conclusions from the experiments, it is important to use reliable optimization methods whose results are as little as possible subject to random disturbances of the solution path.  This is why a widespread numerical optimization procedure with well-defined convergence properties, the conjugate gradient method, has been used in addition the optimization methods usual in the neural network community.

The problems were generated so that they have a known MSE minimum equal to zero (see Sect.~\ref{Sec.TestProblems}), attainable by a particular network architecture (shallow or deep). 
Since it is hardly possible to generate over-determined problems that have a zero minimum for both a shallow and a deep network, cross-validation has been used. 
A pair of zero minimum problems has been generated, one ($ P_s $) with a zero minimum for shallow network $ N_s $, and another ($ P_d $) with a zero minimum for deep network $ N_d $. 
Both network architectures have the same dimensions of input and output vectors, and hidden layer sizes such that the overall numbers of network parameters are close to each other  (the completely identical parameter numbers are difficult to reach).

Every of both problems $ P_s $ and $ P_d $ has been fitted by shallow network $ N_s $ and deep network $ N_d $. 
Comparing the minima reached allows conjectures about the attainable representational capacity of shallow and deep networks.

The scope of this study is only fully connected networks. There seems to be no doubt that specific deep architectures such as those using convolution networks (mimicking spatial operators in image processing) are optimal for specific problems. So, a comparison of deep and shallow networks with such special architectures would make sense only in such specific application settings. 

\section{\uppercase{Test Problems with a Known Minimum}}\label{sec:test-problems}
\label{Sec.TestProblems}
\noindent To assess the performance of training algorithms, it is desirable to use test problems with a known optimum. A construction method is presented in this section. Many statements about the performance of deep learning are based on computing experience with practical problems from various application domains. In the very most cases, the problem is fitting the deep network to real data. This implies that the real minimum is not known---the size of real problems makes it impossible to figure out. This may distort the performance evaluation.  Reaching ``acceptable'' or even ``the best known'' results from the application problem view lets the unanswered question how far we are from the real optimum. It cannot be excluded that all methods find solutions far away from the optimum, and the statements about the algorithms are to a large degree arbitrary. To clarify this aspect, this investigation will make use of problems with a known minimum. Such problems can be generated in the following way:

A network with a set of arbitrary (e.g., random) weights $ w_0 $ is generated. 
Furthermore, a set of random input vectors $ U $ for a mapping to be fitted is produced. 
These inputs, applied to the network $f \left( u, w \right) $, result in a set of output vectors $ Y $.

\begin{equation}
    y_i = f \left( u_i , w_0 \right), \quad i = 1, \dots, n
\end{equation}

The pairs 
\begin{equation}
    \left( u_i, y_i \right), \quad i = 1, \dots, n
\end{equation}

constitute the training set to be fitted.
The least squares objective function
\begin{equation}
    E = \sum_{i=1}^{n}{ \left( y_i - f \left( u_i, w \right) \right)^2 }
\end{equation}

has an obvious minimum of zero. This minimum may not be unique in terms of the parameter vector $ w $. So, the success of the fitting is measured only via a minimum value of $ E $ reached. The random weights are generated from a uniform distribution 

\begin{equation}
    w_i \in \left( \frac{- w_f}{\sqrt{n+1}}, \frac{w_f}{\sqrt{n+1}} \right), \qquad i = 1, \dots, n
\end{equation}

with $ n $ being the number of unit inputs from the preceding layer. There is a predefined factor $ w_f $ for controlling the degree of saturation within the network. The division by the square root of $ n + 1 $ has the goal of reaching an identical standard deviation of the weighted sum (including the bias) going as input to the nonlinear unit.

\section{\uppercase{Optimization Methods}}\label{sec:optimiziation}
\noindent Neural networks were optimized by several methods implemented in the popular framework \emph{Keras} \cite{chollet_Keras_2015a} with the TensorFlow backend\footnote{The actual version is \texttt{2.0.0-beta0}} \cite{tensorflow2015-whitepaper}:
Stochastic Gradient Descent (SGD) and RMSprop were selected because of their widespread use, as well as Adadelta \cite{zeiler_ADADELTAAdaptiveLearningRateMethod_2012}. 

These methods are first-order and there is a widespread opinion in the neural network community that second-order methods are not superior to the first-order ones. However, there are strong theoretical and empirical arguments in favor of the second order methods from numerical mathematics. To make sure that the results in favor of shallow or deep networks are not biased by deficiencies of the optimization methods used, second-order methods should not be neglected. So, the Conjugate Gradient method (CG, see, e.g.~\cite{press_NumericalRecipes_1989}), as implemented in SciPy, has also been included. 
This implementation uses the line search method based on the step length conditions of Wolfe~\cite{wolfe_ConvergenceConditionsAscentMethods_1969}. 
It exploits the derivative information and has excellent convergence properties for smooth functions.

Since the Keras-to-SciPy-to-Keras interface requires a custom built bridge for information flow between the frameworks, GPU-enabled fast executions are not available.

The performance of the optimization methods has been compared by the number of gradient calls (\emph{epochs} in Keras). All methods have been used with Keras' and SciPy's default settings.

\section{\uppercase{Computing Results}}
\noindent A series of computing experiments have been carried out to assess the relationship between attainable representational capacities of shallow and deep networks. The comparison is by using identical mapping problems (defined by input/output pairs) and observing the errors of both network architectures. To provide a reasonable meaning to the mean square figures attained and to make the results comparable, all problems have been deliberately defined to have a minimum at zero, according to the scheme of Section~\ref{Sec.TestProblems}.To justify the use of the hidden layer as a feature extractor, its width should be smaller than the minimum of the input and output sizes. The dimensions have been chosen so that the full regression is under-determined (as typical for the application class mentioned above), but the relatively narrow hidden layer makes it slightly over-determined. So the effect of overfitting, harmful for generalization, is excluded.

Three problem sizes denoted as $ A $, $ B $ and $ C $ have been used. These classes are characterized by their input and output dimensions as well as by the size of the training set. For every class, a shallow network with a single hidden layer and two deep networks with three and five hidden layers have been generated. The problem of size class $ X \in \left\{ A, B, C \right\} $ with $ i $ hidden layers is denoted by $ X_i $. The concrete network sizes, parameter numbers, and numbers of constraints are given in Table~\ref{Tab.ProblemSize}. The numbers of constraints are imposed by the reference outputs to be fitted. It is the product of the output dimension and the training set size. Comparing the number of constraints with the number of parameters defines the extent of over-determination or under-determination of the problem (e.g., a problem with more constraints than parameters is over-determined).

\begin{table*}
	\caption{Overview of test problems.}
	\label{Tab.ProblemSize}
	\centering
	\small
	\begin{tabular}{c rrr rr rr}
		\toprule
		Problem & Input      &    Output & Data size & \# hidden layers &  \# nodes & \# parameters & \# constraints \\ \midrule
		 $A_1$  & \num{ 100} & \num{ 50} & \num{ 80} & \num{1}          & \num{ 20} & \num{  3070}  &   \num{  4000} \\
		 $A_3$  & \num{ 100} & \num{ 50} & \num{ 80} & \num{3}          & \num{ 16} & \num{  3010}  &   \num{  4000} \\
		 $A_5$  & \num{ 100} & \num{ 50} & \num{ 80} & \num{5}          & \num{ 14} & \num{  3004}  &   \num{  4000} \\ \midrule
		 $B_1$  & \num{ 300} & \num{150} & \num{240} & \num{1}          & \num{ 60} & \num{ 27210}  &   \num{ 36000} \\
		 $B_3$  & \num{ 300} & \num{150} & \num{240} & \num{3}          & \num{ 49} & \num{ 27149}  &   \num{ 36000} \\
		 $B_5$  & \num{ 300} & \num{150} & \num{240} & \num{5}          & \num{ 43} & \num{ 27111}  &   \num{ 36000} \\ \midrule
		 $C_1$  & \num{1000} & \num{500} & \num{800} & \num{1}          & \num{200} & \num{300700}  &   \num{400000} \\
		 $C_3$  & \num{1000} & \num{500} & \num{800} & \num{3}          & \num{164} & \num{300784}  &   \num{400000} \\
		 $C_5$  & \num{1000} & \num{500} & \num{800} & \num{5}          & \num{144} & \num{300164}  &   \num{400000} \\ \bottomrule
	\end{tabular}
\end{table*}

Hidden layer units are symmetric sigmoid functions rescaled to have a unity derivative at $ x = 0 $, defined by
\begin{equation}
  s \left( x \right) = \frac{1}{1 + e^{- x}}
\end{equation}
and
\begin{equation}
\label{Eq.Sig2}
  f \left( x \right) = 2 s \left( 2 x \right) - 1 = \frac{2}{1 + e^{-2 x}} - 1
\end{equation}

For every individual network architecture, fifteen different random parametrizations with corresponding training sets have been generated, all with a known mean square error minimum of zero. 

The results for the optimization methods for problem size class $ B $ are given in Table~\ref{Tab.ProblemsSizeB}. The focus of this table is on showing the performance on shallow and deep networks for all optimization methods. Besides to the optimum of the error function $ F_{opt} $, the error value at the initial random parameter set is shown to illustrate the extent of the error improvement. The number of iterations is fixed to \num{2000} for Keras-methods while it is determined by the stopping rule for the conjugate gradient, although the maximum number of iterations for the conjugate gradient is also set to \num{2000}.

The first three blocks of the table show the optimization results for shallow and deep networks individually. Each network is optimized to fit the training set generated for this network architecture. The error optimum is known to be zero by virtue of the principles of Section~\ref{Sec.TestProblems}. These optima set the baseline for those reached by the cross checks in the following rows.

The following four blocks represent the cross check itself. The column \emph{Network} denotes the network architecture used for the applied optimization. The column \emph{Data Source} points to the architecture for which the training set has been generated, with a known error optimum of zero. For the training runs presented in these rows, the optimum is not known. It is only known that it would be zero for the architecture given in the column \emph{Data Source}, but not necessarily also for the architecture of the column \emph{Network}, used for the fitting. 

For example, for the first of these sixteen rows, the network architecture trained is $ B_1 $ (i.e., a shallow net with a single hidden layer). It is optimized to fit the training set for which it is known that a zero error can be reached by the architecture $ B_3 $ (i.e., a deep net with three hidden layers).

The column \emph{Ratio to CG} displays how many times the error function value attained by the Keras methods was higher than that reached by the conjugate gradient.

The column \emph{Deep/Shallow} shows the ratio of the following error function values for the deep network and the shallow network.

\begin{table*}
\caption{Detailed results for problems of size $ B $.}
\label{Tab.ProblemsSizeB}
\centering
\small
\begin{tabular}{cc l rrr rr}
	\toprule
	       Network         &      Data Source       & Algorithm & \# iterations & $ F_{init} \times 10^{-3} $ & $ F_{opt} \times 10^{-3} $ & Ratio to CG   & Deep/Shallow \\ \midrule
	\multirow{4}{*}{$B_1$} & \multirow{4}{*}{$B_1$} & Adadelta  & \num{2000}    &                 \num{332.2} &              \num{  6.748} & \num{ 578.43} &           -- \\
	          -            &                        & RMSprop   & \num{2000}    &                 \num{332.2} &              \num{  0.098} & \num{   8.40} &           -- \\
	                       &                        & SGD       & \num{2000}    &                 \num{332.2} &              \num{ 90.402} & \num{7748.77} &           -- \\
	                       &                        & CG        & \num{ 821}    &                 \num{332.2} &              \num{  0.012} & --            &           -- \\ \midrule
	\multirow{4}{*}{$B_3$} & \multirow{4}{*}{$B_3$} & Adadelta  & \num{2000}    &                 \num{143.3} &              \num{  6.446} & \num{ 173.67} &           -- \\
	                       &                        & RMSprop   & \num{2000}    &                 \num{143.3} &              \num{  0.243} & \num{   6.54} &           -- \\
	                       &                        & SGD       & \num{2000}    &                 \num{143.3} &              \num{ 50.485} & \num{1360.22} &           -- \\
	                       &                        & CG        & \num{2200}    &                 \num{143.3} &              \num{  0.037} & --            &           -- \\ \midrule
	\multirow{4}{*}{$B_5$} & \multirow{4}{*}{$B_5$} & Adadelta  & \num{2000}    &                 \num{ 83.8} &              \num{  4.915} & \num{  41.04} &           -- \\
	                       &                        & RMSprop   & \num{2000}    &                 \num{ 83.8} &              \num{  0.277} & \num{   2.31} &           -- \\
	                       &                        & SGD       & \num{2000}    &                 \num{ 83.8} &              \num{ 33.233} & \num{ 277.51} &           -- \\
	                       &                        & CG        & \num{1490}    &                 \num{ 83.8} &              \num{  0.120} & --            &           -- \\ \midrule
	\multirow{4}{*}{$B_1$} & \multirow{4}{*}{$B_3$} & Adadelta  & \num{2000}    &                 \num{235.8} &              \num{  2.577} & \num{  70.41} &           -- \\
	                       &                        & RMSprop   & \num{2000}    &                 \num{235.8} &              \num{  0.075} & \num{   2.04} &           -- \\
	                       &                        & SGD       & \num{2000}    &                 \num{235.8} &              \num{ 45.396} & \num{1240.02} &           -- \\
	                       &                        & CG        & \num{ 420}    &                 \num{235.8} &              \num{  0.037} & --            &           -- \\ \midrule
	\multirow{4}{*}{$B_1$} & \multirow{4}{*}{$B_5$} & Adadelta  & \num{2000}    &                 \num{206.7} &              \num{  1.594} & \num{  52.44} &           -- \\
	                       &                        & RMSprop   & \num{2000}    &                 \num{206.7} &              \num{  0.070} & \num{   2.29} &           -- \\
	                       &                        & SGD       & \num{2000}    &                 \num{206.7} &              \num{ 32.404} & \num{1065.83} &           -- \\
	                       &                        & CG        & \num{ 333}    &                 \num{206.7} &              \num{  0.030} & --            &           -- \\ \midrule
	\multirow{4}{*}{$B_3$} & \multirow{4}{*}{$B_1$} & Adadelta  & \num{2000}    &                 \num{237.8} &              \num{ 28.331} & \num{   6.75} &  \num{ 11.0} \\
	                       &                        & RMSprop   & \num{2000}    &                 \num{237.8} &              \num{  4.415} & \num{   1.05} &  \num{ 59.2} \\
	                       &                        & SGD       & \num{2000}    &                 \num{237.8} &              \num{118.896} & \num{  28.34} &  \num{  2.6} \\
	                       &                        & CG        & \num{1072}    &                 \num{237.8} &              \num{  4.195} & --            &  \num{114.6} \\ \midrule
	\multirow{4}{*}{$B_5$} & \multirow{4}{*}{$B_1$} & Adadelta  & \num{2000}    &                 \num{208.6} &              \num{ 44.980} & \num{   5.32} &  \num{ 28.2} \\
	                       &                        & RMSprop   & \num{2000}    &                 \num{208.6} &              \num{  9.629} & \num{   1.14} &  \num{138.1} \\
	                       &                        & SGD       & \num{2000}    &                 \num{208.6} &              \num{136.118} & \num{  16.11} &  \num{  4.2} \\
	                       &                        & CG        & \num{2125}    &                 \num{208.6} &              \num{  8.451} & --            &  \num{278.0} \\ \bottomrule
\end{tabular}
\end{table*}

Additionally, Table~\ref{Tab.ProblemsAll} shows mean square minima for all size classes using the Keras optimization method \emph{RMSprop}. This table elucidates the development of the performance (MSE) with shallow and deep networks for varying network sizes. Each row shows the performance of a pair of a shallow and a deep network with a comparable number of parameters. The average performance of a shallow network for a problem for which a zero error minimum is known to be attainable by a deep network is given in the column \emph{Data deep -- NN shallow}. The average performance of a deep network for a problem for which a zero error minimum is known to be attainable by a shallow network is given in the column \emph{Data shallow -- NN deep}. The ratio of both average performances is shown in the column \emph{Ratio Deep/Shallow}.

\begin{table*}
\caption{Results for all given problems and their ratio between shallow and deep networks.}\label{Tab.ProblemsAll}
\centering
\small
\begin{tabular}{cc rr r}
	\toprule
	       Shallow         &  Deep   & Data deep -- NN shallow {\scriptsize $ \times 10^{-3} $} & Data shallow -- NN deep {\scriptsize $ \times 10^{-3} $} & Ratio Deep/Shallow \\ \midrule
	\multirow{2}{*}{$A_1$} & $ A_3 $ & \num{0.038}                                              &                                             \num{ 5.368} & \num{140.5}        \\
	                       & $ A_5 $ & \num{0.035}                                              &                                             \num{11.353} & \num{320.5}        \\
	\multirow{2}{*}{$B_1$} & $ B_3 $ & \num{0.075}                                              &                                             \num{ 4.415} & \num{ 59.2}        \\
	                       & $ B_5 $ & \num{0.070}                                              &                                             \num{ 9.629} & \num{138.1}        \\
	\multirow{2}{*}{$C_1$} & $ C_3 $ & \num{0.182}                                              &                                             \num{ 4.663} & \num{ 25.7}        \\
	                       & $ C_5 $ & \num{0.148}                                              &                                             \num{ 9.777} & \num{ 66.1}        \\ \bottomrule
\end{tabular}
\end{table*}

The following can be observed:
\begin{itemize}
    \item The mean square error minima (MSE) attained by shallow networks for problems having a zero MSE for some deep network are essentially lower than in the opposite situation.
    \item The difference tends to slightly decrease with the problem size.
    \item The by far weakest method was the SGD, while the best was the conjugate gradient (CG). Adadelta and RMSprop were performing between the both, with RMSprop sometimes approaching the CG performance.
    \item The difference between the performance with a shallow network on one hand and deep network on the other hand grows with the performance of the optimization method: the difference is relatively small for the worst performing SGD and very large for the best performing CG.
\end{itemize}

\section{\uppercase{Discussion}}\label{sec:discussion}
\noindent The computing experiments seem to essentially show the superiority of shallow networks in attaining low mean square minima for given mapping problems. This is not necessarily a contradiction to the theoretical results expecting the contrary. It is still possible that the representational capacity of deep networks is superior, while it is difficult to exploit this capacity by fitting the mapping with help of numerical algorithms.

Shallow networks have been superior for all test problems and all optimizing algorithms. However, it is interesting to observe that the gap, although always large, was relatively smaller for weakly performing optimization methods (SGD and Adadelta) as well as for large networks.

A possible hypothesis summarizing the both might be that the gap is low if the optimizing method fails to search for the minimum efficiently, approaching the performance of some kind of random search. This can result either from the weakness of the method itself or from the difficulty of the problem. It is clear that even sophisticated methods such as the conjugate gradient have growing difficulties with growing problem size. These difficulties may have to do with the machine precision necessary for stopping rules (testing for \emph{zero gradient}) or with the number of iterations available. 

So, the conjugate gradient provides a theoretical guarantee for finding a minimum for an exactly quadratic problem of dimension $ q $ in $ q $ steps. This is obviously a huge number of iterations for our test problems (and other real-world ones). In addition to this, our problems are far from being exactly quadratic (they may even be non-convex), which further increases the computing requirements. This makes clear that the adequacy of every optimization method decreases with the problem size. This still does not explain why deep networks should be more favorable if the optimization method is not adequate to the problem---at best, it may be argued that the search is then close to the random search, which might be indifferent to the functional parametrization optimized.

\bibliographystyle{apalike}
{\small \bibliography{references}}

\begin{thebibliography}{}

\bibitem[Abadi et~al., 2015]{tensorflow2015-whitepaper}
Abadi, M., Agarwal, A., Barham, P., Brevdo, E., Chen, Z., Citro, C., Corrado,
  G.~S., Davis, A., Dean, J., Devin, M., Ghemawat, S., Goodfellow, I., Harp,
  A., Irving, G., Isard, M., Jia, Y., Jozefowicz, R., Kaiser, L., Kudlur, M.,
  Levenberg, J., Mané, D., Monga, R., Moore, S., Murray, D., Olah, C.,
  Schuster, M., Shlens, J., Steiner, B., Sutskever, I., Talwar, K., Tucker, P.,
  Vanhoucke, V., Vasudevan, V., Viégas, F., Vinyals, O., Warden, P.,
  Wattenberg, M., Wicke, M., Yu, Y., and Zheng, X. (2015).
\newblock {{TensorFlow}}: {{Large}}-{{Scale Machine Learning}} on
  {{Heterogeneous Systems}}.
\newblock Software available from https://tensorflow.org.

\bibitem[Bengio et~al., 2004]{bengio_OutofsampleExtensionsLleIsomapMds_2004}
Bengio, Y., Paiement, J.-f., Vincent, P., Delalleau, O., Roux, N.~L., and
  Ouimet, M. (2004).
\newblock Out-of-sample extensions for lle, isomap, mds, eigenmaps, and
  spectral clustering.
\newblock In {\em Advances in Neural Information Processing Systems}, pages
  177--184.

\bibitem[Chollet et~al., 2015]{chollet_Keras_2015a}
Chollet, F. et~al. (2015).
\newblock Keras.
\newblock Software available from https://keras.io.

\bibitem[Erhan et~al.,
  2009]{erhan_DifficultyTrainingDeepArchitecturesEffect_2009}
Erhan, D., Manzagol, P.-A., Bengio, Y., Bengio, S., and Vincent, P. (2009).
\newblock The difficulty of training deep architectures and the effect of
  unsupervised pre-training.
\newblock In {\em Artificial {{Intelligence}} and {{Statistics}}}, pages
  153--160.

\bibitem[Goodfellow et~al.,
  2013]{goodfellow_MultidigitNumberRecognitionStreetView_2013}
Goodfellow, I.~J., Bulatov, Y., Ibarz, J., Arnoud, S., and Shet, V. (2013).
\newblock Multi-digit number recognition from street view imagery using deep
  convolutional neural networks.
\newblock {\em arXiv preprint arXiv:1312.6082}.

\bibitem[LeCun et~al., 2015]{lecun_DeepLearning_2015}
LeCun, Y., Bengio, Y., and Hinton, G. (2015).
\newblock Deep learning.
\newblock {\em Nature}, 521(7553):436--444.

\bibitem[Montufar et~al., 2014]{montufar_NumberLinearRegionsDeepNeural_2014}
Montufar, G.~F., Pascanu, R., Cho, K., and Bengio, Y. (2014).
\newblock On the {{Number}} of {{Linear Regions}} of {{Deep Neural Networks}}.
\newblock In Ghahramani, Z., Welling, M., Cortes, C., Lawrence, N.~D., and
  Weinberger, K.~Q., editors, {\em Advances in {{Neural Information Processing
  Systems}} 27}, pages 2924--2932. {Curran Associates, Inc.}

\bibitem[Press et~al., 1989]{press_NumericalRecipes_1989}
Press, W.~H., Flannery, B.~P., Teukolsky, S.~A., Vetterling, W.~T., et~al.
  (1989).
\newblock {\em Numerical Recipes}, volume~2.
\newblock {Cambridge university press Cambridge}.

\bibitem[Wolfe, 1969]{wolfe_ConvergenceConditionsAscentMethods_1969}
Wolfe, P. (1969).
\newblock Convergence conditions for ascent methods.
\newblock {\em SIAM review}, 11(2):226--235.

\bibitem[Zeiler, 2012]{zeiler_ADADELTAAdaptiveLearningRateMethod_2012}
Zeiler, M.~D. (2012).
\newblock {{ADADELTA}}: {{An Adaptive Learning Rate Method}}.
\newblock {\em arXiv:1212.5701 [cs]}.

\end{thebibliography}

\end{document}